\documentclass[11pt]{article}
\usepackage[final]{acl}

\usepackage{times}
\usepackage{latexsym}
\usepackage[T1]{fontenc}
\usepackage[utf8]{inputenc}
\usepackage{microtype}
\usepackage{graphicx}
\usepackage{booktabs}
\usepackage{amsmath,amssymb}
\usepackage{xcolor}
\usepackage{algorithm}
\usepackage{algpseudocode}
\usepackage{subcaption}
\usepackage{enumitem}
\definecolor{basegrey}{HTML}{8A8A8A}
\definecolor{elite}{HTML}{E8417A}
\newcommand{\method}{\textsc{EliteReplay}}

\usepackage{algorithm}
\usepackage{algpseudocode}
\usepackage{booktabs}
\usepackage{amsmath}
\usepackage{xcolor}
\usepackage[most]{tcolorbox}
\usepackage{listings}
\newif\ifshowcomments
\showcommentstrue

\title{\method{}: Preserving Diverse Solutions in RL Post-Training\\ via Quality-Diversity Search}

\author{Hanane Nour Moussa \quad Doeun Lee \quad Sachin Kumar \\
  The Ohio State University \\
  \texttt{\{moussa.45, lee.11501, kumar.1145\}@osu.edu}}

\begin{document}
\maketitle

\begin{abstract}
Reinforcement learning has become a common post-training technique to improve the accuracy of LLMs on verifiable tasks. However, such training can narrow the set of solutions the model can produce. Several approaches have attempted to address this behavior-sharpening effect, but they often optimize metrics that do not reflect the diversity of \emph{correct} solutions. Ideally, we want a model that can produce multiple good solutions, not only a single reliable one. We propose \method{}, a post-training recipe that pairs GRPO with an ``elite'' archive built by quality-diversity search. Before training, we use the frozen base model to collect solutions that span distinct approaches; during training, GRPO is combined with a replay loss on this archive. The result is a post-trained model that improves accuracy while preserving more of the base model's solution diversity. On multiple text-to-SQL tasks, \method{} achieves the highest pass@$128$ than all baselines, producing up to 77\% more distinct correct strategies than the strongest baselines. It preserves most of the base model's correct strategies, outperforming existing methods in strategy retention. This diversity translates into practical benefits on ambiguous questions, where \method{} covers both interpretations nearly twice as often as the strongest baseline on AmbiQT\footnote{All artifacts can be found at \url{https://github.com/skai-research/EliteReplay}}.
\end{abstract}

\section{Introduction}
\label{sec:intro}
Reinforcement learning with verifiable rewards (RLVR) has become a standard post-training technique for large language model (LLM) reasoning \citep{shao2024deepseekmath,guo2025deepseekr1}. However, studies show that common RLVR methods such as Group Relative Policy Optimization (GRPO) improve performance via \emph{sharpening}, i.e. concentrating probability on a few behaviors the base model already had \citep{yue2025does,zhao2025echo,huang2025sharpening}. RL post-training therefore improves \emph{reliability} (pass@$1$), but often by reducing \emph{reach} (pass@$k$) \citep{oh2026sharpening}.

\begin{figure}[t]
\centering
\includegraphics[width=0.95\columnwidth]{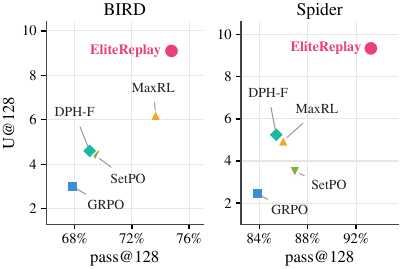}
\caption{\textbf{Accuracy versus strategy diversity at $k=128$.} Each point is a post-trained model on BIRD (in-distribution) and Spider (out-of-distribution). \method{} reaches the highest pass@128 while producing far more distinct correct strategies.}
\label{fig:teaser}
\end{figure}%
Several works seek to recover this lost reach by improving pass@$k$ relative to the base model \citep{tajwar2026maxrl,li2025dphrl,li2026setpo,walder2025pkpo}. Since recovering reach may require retaining alternative ways of solving problems, pass@$k$ has also become a common proxy for diversity \citep{dang2025weight,song2025outcome,chen2025passk,yu2025passk,jiang2026mentor}. Yet pass@$k$ only asks whether some sample is correct, not whether the correct samples differ. A model can therefore achieve high pass@$k$ by reliably producing a single strategy for each question while losing the alternatives the base model once exhibited. 

Retaining these alternative ways of solving a problem, its solution diversity, is important whenever the approach matters alongside the final answer. Different correct solutions may follow distinct reasoning paths, differ in computational efficiency, or encode different interpretations of an ambiguous request. Systems that rely on repeated sampling and selection also benefit from having meaningful variety \citep{li2022alphacode,pourreza2025chase}. As Figure~\ref{fig:teaser} shows, methods with nearly identical pass@128 can differ substantially in the number of distinct correct strategies they produce.

Preserving a diverse set of good solutions is the central concern of \emph{quality-diversity} (QD) optimization. Rather than seeking a single optimum, QD algorithms such as novelty search \citep{lehman2011abandoning} and MAP-Elites \citep{mouret2015illuminating} maintain an archive of high-performing solutions spread across a space of behavior descriptors. 
In the deep RL literature, such archives have been coupled with policy-gradient learners to broaden exploration, enabling the discovery of new behaviors in policies trained from scratch \citep{khadka2018evolution,pourchot2019cemrl,ecoffet2021first}. Post-training a language model, however, presents a different challenge, since the base model already exhibits a variety of correct strategies that RL can suppress. At the same time, language models offer a natural advantage for QD search, as the base model itself can propose meaningful variations through repeated sampling \citep{lehman2023elm,romera2024funsearch,bradley2024qdaif,samvelyan2024rainbow}.

We study whether quality-diversity archives can preserve solution diversity during GRPO training. Our method, \method{}, first runs MAP-Elites with the frozen base model to build an \emph{elite archive}: a set of verified solutions, with at most one solution per cell of a grid indexed by a behavior descriptor. It then trains with GRPO while adding a small replay loss on this archive.

Measuring diversity requires distinguishing how solutions are obtained, not just whether they are correct. We therefore evaluate diversity at the level of solution \emph{strategies}, using a structural description of how each solution solves the question, computed deterministically from the solution itself. This allows us to measure U@$k$, the number of unique correct strategies among $k$ samples, alongside effective diversity \citep{shypula2025evaluating}, the probability that two random samples are both correct and different, which captures improvements in reliability without sacrificing diversity.

We evaluate \method{} on text-to-SQL, a natural testbed where questions can admit multiple correct queries and natural-language ambiguity is common. As shown in Figure~\ref{fig:teaser}, \method{} achieves 74.8\% pass@$128$ on BIRD \citep{li2023bird}, comparable to MaxRL's 73.7\%, while producing 9.1 distinct correct strategies per question compared to just 6.2. On the out-of-distribution Spider \citep{yu2018spider}, it achieves 93.2\% pass@$128$, outperforming the strongest baseline by 6.3 percentage points while producing up to 77\% more distinct correct strategies. \method{} also retains 62\% and 60\% of the base model's correct strategies on BIRD and Spider, respectively, at least 23 percentage points more than any baseline. This preserved diversity also benefits ambiguous questions, with \method{} covering both interpretations on 27.0\% of AmbiQT \citep{bhaskar2023ambiqt} questions, compared to 14.2\% for the strongest baseline.

Our contributions are:
\begin{itemize}
\setlength\itemsep{0pt}
\item We propose \method{}, which anchors RL to an elite archive constructed through quality-diversity search using the base model (\S\ref{sec:method}).
\item We evaluate RL post-training through \emph{strategy-level} diversity, measuring unique correct strategies (U@$k$), effective diversity, and retention of the base model's strategies (\S\ref{sec:eval}).
\item We show that \method{} preserves more diverse correct strategies while maintaining strong pass@$k$, retains most base-model strategies, and better covers ambiguous questions (\S\ref{sec:results}).
\end{itemize}

\begin{figure*}[t]
\centering
\includegraphics[width=\textwidth]{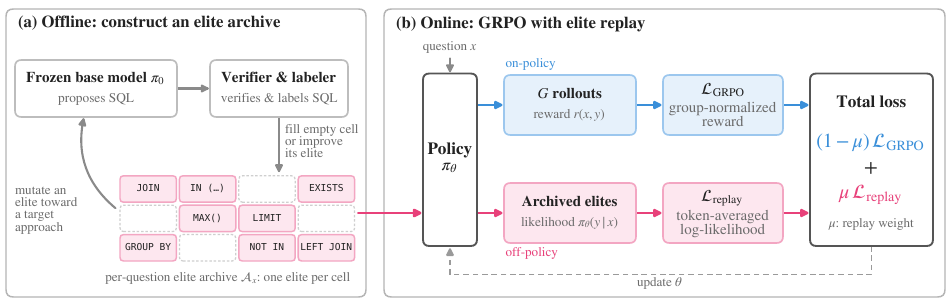}
\caption{\textbf{Overview of \method{}.} \textbf{(a)} Before training, MAP-Elites with the frozen base model builds a per-question elite archive, keeping one verified solution per behavior cell. \textbf{(b)} During training, GRPO is combined with a replay loss on the fixed archive. The figure illustrates the method on text-to-SQL.}
\label{fig:method}
\end{figure*}%

\section{\method{}}
\label{sec:method}

Our proposed approach improves accuracy with RL while preserving the base model's distinct correct solutions in two stages (Figure~\ref{fig:method}). Before training, QD search with the frozen base model builds an \emph{elite archive} of verified solutions that span distinct approaches (\S\ref{sec:method-archive}). During training, GRPO is combined with a replay loss on this archive (\S\ref{sec:method-replay}). Appendix~\ref{app:method_details} details both stages.

\subsection{Constructing the elite archive}
\label{sec:method-archive}
RL post-training tends to concentrate probability on a narrower set of approaches \citep{yue2025does,huang2025sharpening}, putting alternative correct approaches at risk of disappearing. Simply replaying solutions sampled from the base model may not prevent this collapse, since such samples are dominated by the base model's more frequent approaches and underrepresent rarer ones. We therefore build the replay set using quality-diversity search \citep{pugh2016quality}, instantiating MAP-Elites \citep{mouret2015illuminating}, a QD algorithm that partitions a space of behavior descriptors into cells and retains the best solution found in each, on a subset $\mathcal{D}_A$ of the training questions, using the frozen base model $\pi_0$ to generate candidate solutions. For each question $x \in \mathcal{D}_A$, we maintain an archive $\mathcal{A}_x$ of verified solutions, organized into cells by solution approach. Each cell stores at most one solution, or \emph{elite}, allowing the archive to represent multiple correct approaches to the same question.

\paragraph{Defining the behavior space.}
Classical MAP-Elites relies on behavior descriptors designed by hand for each domain. Instead, we first collect a set of verified seed solutions from the base model and provide them to a separate LLM, the \emph{labeler}, as representative examples of how the task can be solved. Using these examples, the labeler induces a broader vocabulary of possible approaches, including ones not observed in the seeds, to guide subsequent exploration. The labeler assigns each solution an approach label, capturing its general method of solving the question. To distinguish solutions that follow the same broad approach but differ in their structure, we combine this label with structural features extracted deterministically from the solution (e.g., joins or subqueries in SQL, loops or recursion in code). The resulting behavior descriptor $b(y)$ determines the solution's archive cell.
This descriptor is used only to guide search; we evaluate diversity with a separate, deterministic notion of strategy (\S\ref{sec:eval}). Appendix~\ref{app:behavior} details the behavior space, and Appendices~\ref{app:sql-archive} and~\ref{app:prompts} give the features and labeler prompts used in our text-to-SQL experiments.
\paragraph{Constructing the archive.}
Given this behavior space, we populate and expand the per-question archives through four steps.
\begin{enumerate}[label=(\arabic*), leftmargin=*, itemsep=0pt, topsep=2pt]
\item \textbf{Initialization.} We insert the verified seed solutions into their corresponding archive cells, establishing the initial set of elites for each question.
\item \textbf{Selection.} We choose a target approach among those the labeler judges applicable to the question, prioritizing approaches not yet represented in the archive while balancing attempts across targets (Appendix~\ref{app:search}). We then select a parent elite, preferably one following a different approach.
\item \textbf{Variation.} We prompt $\pi_0$ with the original question, target approach, and parent elite, asking it to generate a new solution following the target approach.
\item \textbf{Evaluation and insertion.} We verify the candidate and, if correct, assign it to the cell corresponding to its actual approach, which may differ from the requested one. The candidate fills an empty cell or replaces an existing elite if it is shorter.
\end{enumerate}
We repeat selection, variation, and evaluation for a fixed search budget. Because selection favors untried approaches, search is directed toward approaches that may be correct but rarely sampled by the base model, while candidates repeating an existing approach only compete for an occupied cell. Search more than doubles the cells occupied by seed sampling and yields more distinct elites than direct sampling (Appendices~\ref{app:archive-growth} and~\ref{app:iid}).
The surviving elites, paired with their original task prompts, form the replay set.
\subsection{Training with elite replay}
\label{sec:method-replay}

Starting from the base model $\pi_0$, we train the policy $\pi_\theta$ with GRPO \citep{shao2024deepseekmath}, while encouraging it to reproduce solutions from the archive. GRPO improves correctness, and archive replay helps preserve alternative approaches.

\paragraph{Replaying archived solutions.}
At each training step, we sample a batch $\mathcal{E}$ of verified solutions from the archive, balancing sampling across questions and their occupied cells so that questions with more elites do not dominate. To retain these solutions, we define the replay loss as the negative log-likelihood of the archived solutions, normalized by solution length so that longer solutions do not receive more weight,
\begin{equation}
\mathcal{L}_{\mathrm{replay}}(\theta) = -\frac{1}{|\mathcal{E}|}\sum_{(x,y)\in\mathcal{E}} \frac{1}{|y|}\log\pi_\theta(y\mid x).
\end{equation}

\paragraph{Combining replay with GRPO.}
Inspired by CHORD \citep{zhang2025chord}, we combine the on-policy GRPO objective with the supervised replay loss using a simplified formulation with a fixed mixing weight $\mu$ 
\begin{equation}
\mathcal{L}(\theta) = (1-\mu)\,\mathcal{L}_{\mathrm{GRPO}}(\theta) + \mu\,\mathcal{L}_{\mathrm{replay}}(\theta).
\end{equation}
The archive remains fixed, and the replay loss provides a consistent signal for preserving alternative solutions alongside GRPO updates (Appendix~\ref{app:replay}). At inference, the trained policy is sampled directly, without archive lookup or search.

\section{Evaluating Strategy Diversity}
\label{sec:eval}

We measure how well RL post-training preserves the range of correct solutions at the level of solution \emph{strategies}, using three complementary metrics that capture the number of distinct correct strategies, the structural diversity of correct samples, and the retention of strategies from the base model.

\paragraph{Strategies.}
We represent each solution $y$ by a strategy $s(y)$, a deterministic description of how the solution is constructed that abstracts away surface details. For solutions with formal structure, such as code, abstract syntax trees (ASTs) provide a natural basis for distinguishing programs by their structure rather than their surface form \citep{ren2020codebleu,song2024tsed,shypula2025evaluating}. We therefore derive $s(y)$ from the AST of each SQL query, distinguishing, for example, a query that joins two tables from one that filters through a nested subquery (Appendix~\ref{app:descriptors}).

\paragraph{Unique correct strategies (U@$k$).}
Let $r(x,y)\in\{0,1\}$ indicate whether solution $y$ to question $x$ is correct, and let $y_1,\dots,y_n$ be $n$ samples for $x$. U@$k$ is the expected number of distinct strategies among the correct samples in a random subset of $k$ samples. If $c_s$ of the $n$ samples are correct with strategy $s$, then, analogous to the unbiased pass@$k$ estimator \citep{chen2021codex},
\begin{equation}
\mathrm{U}@k(x) = \sum_{s} \Big[\,1 - \tbinom{n - c_s}{k} \big/ \tbinom{n}{k}\Big],
\end{equation}
averaged over questions. Unlike pass@$k$, which counts whether at least one correct solution is found, U@$k$ counts how many different correct strategies are represented. The two coincide when all correct samples for a question share a single strategy, while U@$k$ increases as additional correct strategies are found.
\begin{figure*}[t]
\centering
\includegraphics[width=0.95\textwidth]{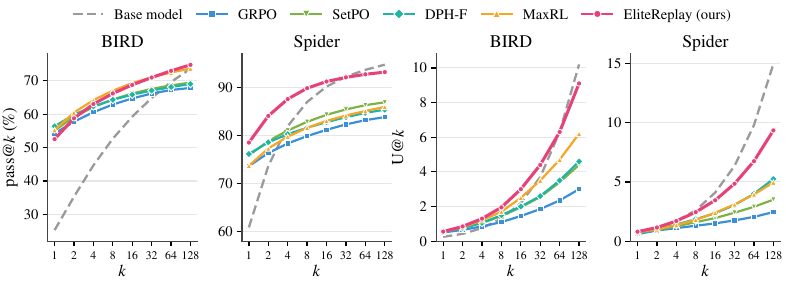}
\caption{\textbf{Pass@$k$ and U@$k$ as the sampling budget grows} on BIRD and Spider. \method{} improves accuracy relative to the base model while producing far more distinct correct strategies than any baseline.}
\label{fig:curves}
\end{figure*}%
\paragraph{Effective diversity.}
Effective diversity \citep{shypula2025evaluating} captures how structurally different correct samples are from one another, measuring the expected dissimilarity between two random samples while counting only pairs in which both are correct,
\begin{equation}
\mathrm{ED}(x) = \frac{2}{n(n-1)} \sum_{i<j} r_i\, r_j \big(1 - \mathrm{sim}(y_i, y_j)\big),
\end{equation}
averaged over questions, where $r_i = r(x,y_i)$ and $\mathrm{sim}\in[0,1]$ is an AST-based similarity. By accounting for both correctness and structural differences, effective diversity rewards models that produce correct solutions reliably without collapsing onto similar outputs.

\paragraph{Strategy retention.}
Let $\mathcal{S}_m(x)$ be the set of strategies that model $m$ produces correctly among its $n$ samples for $x$, and let $\mathcal{Q}$ be the questions on which the base model $\pi_0$ produces at least two distinct correct strategies. Retention measures the fraction of these base-model strategies that remain accessible after post-training,
\begin{equation}
\mathrm{Ret}(m) = \frac{1}{|\mathcal{Q}|} \sum_{x\in\mathcal{Q}} \frac{|\mathcal{S}_m(x) \cap \mathcal{S}_{\pi_0}(x)|}{|\mathcal{S}_{\pi_0}(x)|}.
\end{equation}
\section{Experimental Setup}
\label{sec:setup}
\paragraph{Model and data.}
We use Qwen3.5-4B-Base \citep{qwen35} as our base model and train on 5,000 questions from BIRD \citep{li2023bird}, a text-to-SQL dataset built from real-world databases. We construct the elite archive using a subset of 1,000 of these questions, which remain part of the RL training pool (Appendix~\ref{app:sql-archive}). For evaluation, we use BIRD dev to measure in-distribution performance and Spider 1.0 dev \citep{yu2018spider} to assess out-of-distribution generalization on databases disjoint from those used in training. On both benchmarks, correctness is measured by execution accuracy, comparing the results of generated SQL queries with those of the gold queries on their respective databases (Appendix~\ref{app:verification}).

To examine the post-training effects in an unrelated domain, we also evaluate on MATH-500 \citep{hendrycks2021math}. Finally, to assess the practical value of preserving diverse solutions, we evaluate on AmbiQT \citep{bhaskar2023ambiqt}, a benchmark derived from Spider where each question admits two valid interpretations due to ambiguity (e.g. in column or table names, join paths, etc). We draw $n=128$ samples per question.

\paragraph{Baselines.}
We compare \method{} against standard GRPO \citep{shao2024deepseekmath} and three methods aimed at preserving diversity or improving pass@$k$. \textbf{MaxRL} \citep{tajwar2026maxrl} upweights hard questions through a maximum-likelihood objective, while \textbf{DPH-F} \citep{li2025dphrl} uses forward-KL regularization to preserve the base model's behavior on questions it already solves reliably. \textbf{SetPO} \citep{li2026setpo} encourages diversity by adding to each rollout's advantage a bonus for how much it increases the semantic diversity of its sampled group. All methods are trained on the same data for 100 steps, with \method{} using a replay weight of $\mu=0.05$ (Appendix~\ref{app:mu}). Appendix~\ref{app:compute} gives further implementation and compute details.
\section{Results and Analysis}
\label{sec:results}


\subsection{Accuracy and strategy diversity}
\label{sec:results-coverage}
\begin{figure}[t]
\centering
\includegraphics[width=0.9\columnwidth]{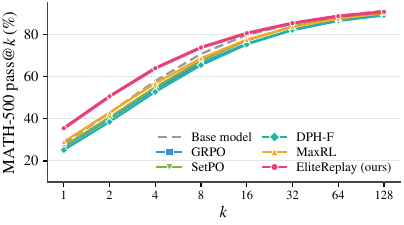}
\caption{\textbf{MATH-500 pass@$k$ of BIRD-trained models.} Baselines barely improve or degrade relative to the base model, whereas \method{} substantially improves it, despite not being trained on math.}
\label{fig:math}
\end{figure}%
\begin{figure}[t]
\centering
\includegraphics[width=0.95\columnwidth]{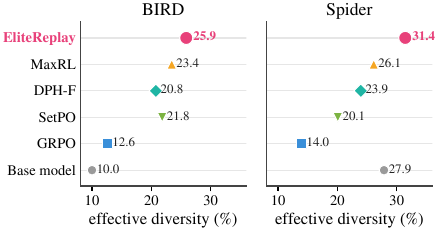}
\caption{\textbf{Effective diversity}, the probability (\%) that two samples are both correct and structurally different. \method{} scores highest on both benchmarks.}
\label{fig:effdiv}
\end{figure}%
\begin{figure}[t]
\centering
\includegraphics[width=0.95\columnwidth]{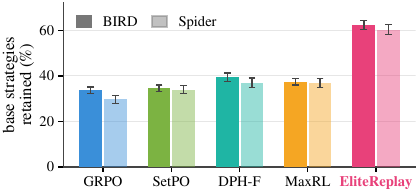}
\caption{\textbf{Share of the base model's correct strategies retained by each method}, over questions on which the base model uses at least two strategies. \method{} has the highest retention rate.}
\label{fig:retention}
\end{figure}%

\paragraph{\method{} improves reliability while substantially preserving diversity.}
All post-trained methods improve pass@$1$ on BIRD, raising it from 25.3\% for the base model to 52--56\%. However, most lose reach as the sampling budget increases (Figure~\ref{fig:curves}), as observed for RLVR on math \citep{yue2025does}. GRPO, SetPO, and DPH-F attain only 68--69\% pass@$128$, below the base model's 73.6\%, whereas MaxRL and \method{} maintain comparable coverage at 73.7\% and 74.8\%, respectively. On out-of-distribution Spider, \method{} achieves the highest pass@$k$ among post-trained methods at every budget, reaching 78.6\% pass@$1$ and 93.2\% pass@$128$, versus at most 76.2\% and 86.9\% for the baselines.

\paragraph{\method{} also maintains performance outside the training domain.}
On MATH-500 (Figure~\ref{fig:math}) three baselines perform worse than the base model across all sampling budgets, with pass@$1$ falling to 25--27\% from 28.3\%. In contrast, \method{} improves pass@$1$ to 35.5\% while remaining within one point of the base model at large sampling budgets. These results suggest that replay may act as a rehearsal mechanism that mitigates out-of-domain forgetting, which is an interesting area of analysis for future work.

\paragraph{Similar pass@$k$ can conceal substantial differences in strategy diversity.}
Despite achieving nearly identical pass@$128$ on BIRD, \method{} produces almost 50\% more distinct correct strategies than MaxRL, with U@$128$ of 9.1 versus 6.2 (Figure~\ref{fig:curves}, right). Thus, MaxRL recovers coverage while relying on fewer ways of solving each question, a distinction that pass@$k$ does not capture. The diversity gap is even larger on Spider, where \method{} produces nearly 80\% more distinct correct strategies than the strongest baseline (9.4 versus 5.3). On BIRD, \method{} also matches or exceeds the base model's U@$k$ up to $k=64$, despite substantially higher pass@$1$, while all baselines fall increasingly behind it at larger budgets.

Effective diversity provides complementary evidence (Figure~\ref{fig:effdiv}). \method{} achieves the highest scores on both benchmarks, reaching 25.9\% on BIRD versus 23.4\% for the strongest baseline. On Spider, it is the only post-training method to exceed the base model (31.4\% versus 27.9\%).

\subsection{Preserving base-model strategies}
\label{sec:results-retention}

\paragraph{\method{} retains substantially more of the base model's correct strategies.}
U@$k$ counts distinct strategies, whereas strategy retention tracks which of the base model’s strategies survive post-training. As shown in Figure~\ref{fig:retention}, the baselines retain only 34--39\% of these strategies on BIRD and 30--37\% on Spider, whereas \method{} retains 62\% and 60\%, respectively.

\paragraph{\method{} preserves broader coverage across individual strategies.}
To examine whether the lost strategies are concentrated in a few approaches or spread across many, we measure how often each of the base model's most common correct strategies are still produced across questions, relative to its original coverage (Appendix Figure~\ref{fig:delta}; details in Appendix~\ref{app:retention}). The baselines lose coverage on nearly every strategy, with reductions of up to 28 percentage points on BIRD. In contrast, \method{} shows smaller reductions on every strategy, never exceeding 9 percentage points. The same pattern holds on Spider (Appendix~\ref{app:delta-spider}).

\paragraph{The baselines often collapse onto a single strategy, while \method{} preserves alternatives.}
To examine how strategy loss manifests at the question level, we compare the distribution of correct samples across strategies for three representative BIRD questions (Figure~\ref{fig:cases}; details in Appendix~\ref{app:cases}). While the baselines often concentrate nearly all correct samples on a single approach, \method{} maintains multiple strategies closer to the base model's original distribution. Notably, the baselines sometimes converge on an approach that the base model rarely used, showing that strategy collapse does not necessarily preserve the base model's preferred behavior, as \citet{zhao2025echo} also observe for RL on math. Spider examples and representative queries for each strategy are provided in Appendices~\ref{app:cases-spider} and~\ref{app:queries}.

\begin{figure*}[t]
\centering
\includegraphics[width=0.95\textwidth]{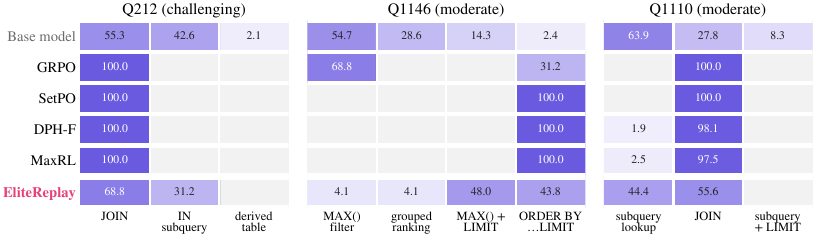}
\caption{\textbf{How correct samples split across strategies on three BIRD questions.} Each row shows the percentage of a method's correct samples (out of 128 rollouts) that use each strategy. The baselines concentrate on one strategy, while \method{} keeps several.}
\label{fig:cases}
\end{figure*}%

\subsection{Diversity under ambiguity}
\label{sec:results-ambiqt}

\paragraph{Preserved diversity improves coverage of ambiguous questions.}
Ambiguous questions provide a practical test of solution diversity, since covering multiple interpretations requires producing distinct correct solutions. For example, ``employees of a department'' could mean current employees or everyone ever assigned to it. A model that produces only one interpretation cannot cover both, however reliably it generates that answer. On AmbiQT, we report Both@$k$, the fraction of questions for which $k$ samples match the gold execution result of each interpretation (Table~\ref{tab:ambiqt}; Appendix~\ref{app:ambiqt}).

\begin{table}[t]
\centering
\small
\setlength{\tabcolsep}{3.4pt}
\begin{tabular}{l ccccccc}
\toprule
$k$ & 2 & 4 & 8 & 16 & 32 & 64 & 128 \\
\midrule
\textcolor{basegrey}{\textit{Base model}} & \textcolor{basegrey}{2.3} & \textcolor{basegrey}{5.5} & \textcolor{basegrey}{10.3} & \textcolor{basegrey}{16.3} & \textcolor{basegrey}{23.0} & \textcolor{basegrey}{30.1} & \textcolor{basegrey}{37.0} \\
\midrule
GRPO & 1.6 & 2.0 & 2.4 & 3.0 & 3.5 & 3.9 & 4.2 \\
SetPO & 2.5 & 5.0 & 5.5 & 6.7 & 7.8 & 8.8 & 10.0 \\
DPH-F & 2.4 & 4.1 & 5.9 & 7.6 & 9.3 & 10.8 & 12.0 \\
MaxRL & 1.9 & 3.7 & 6.0 & 8.5 & 10.8 & 12.7 & 14.2 \\
\textbf{\method{}} & \textbf{3.0} & \textbf{5.9} & \textbf{9.5} & \textbf{13.5} & \textbf{17.9} & \textbf{22.5} & \textbf{27.0} \\
\bottomrule
\end{tabular}
\caption{\textbf{Both@$k$ on AmbiQT (\%)}, the share of questions for which $k$ samples include correct queries for both valid readings. \method{} covers both readings most often among post-trained models.}
\label{tab:ambiqt}
\end{table}

At $k=128$, GRPO covers both interpretations on only 4.2\% of questions, while the strongest diversity-preserving baseline reaches 14.2\%. \method{} nearly doubles this coverage, achieving 27.0\%, with its advantage growing as the sampling budget increases. This improvement closes more than half the gap between the strongest baseline and the base model, which achieves 37.0\%. 

\subsection{Ablating archive construction and replay}
\label{sec:results-archive}

We isolate the contributions of archive construction and replay through three comparisons, using as a control an independently and identically distributed (IID) archive of verified base-model samples at equal generation budget (Appendix~\ref{app:iid}). 
\begin{figure}[t]
\centering
\includegraphics[width=0.9\columnwidth]{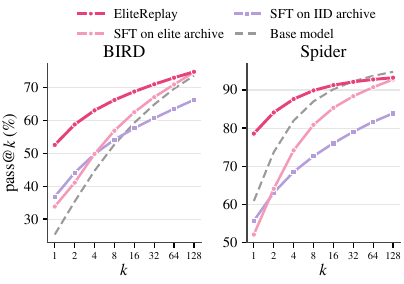}
\caption{\textbf{SFT on the archive alone.} SFT on the elite archive beats the IID archive at large $k$, but neither reaches \method{}'s pass@1.}
\label{fig:sft}
\end{figure}%

\paragraph{Quality-diversity search produces a more effective training archive than IID sampling.}
To isolate the effect of archive composition, we first apply supervised fine-tuning (SFT) to the base model on each archive without RL. SFT on the elite archive achieves 74.6\% pass@$128$ on BIRD and 92.8\% on Spider, outperforming SFT on the IID archive by 8--9 percentage points, with larger gains at higher sampling budgets (Figure~\ref{fig:sft}). However, neither archive alone matches the reliability of \method{}, reaching at most 37\% pass@$1$ on BIRD compared to 52.5\% for \method{}. This shows that the archive improves coverage, while RL provides the substantial gains in low-sample accuracy.

\begin{table}[t]
\centering
\small
\setlength{\tabcolsep}{4pt}
\begin{tabular}{l cc cc}
\toprule
& \multicolumn{2}{c}{pass@$k$ (\%)} & \multicolumn{2}{c}{U@$k$} \\
\cmidrule(lr){2-3}\cmidrule(lr){4-5}
Variant & $k=1$ & $k=128$ & $k=16$ & $k=128$ \\
\midrule
\method{} & 52.5 & 74.8 & \textbf{3.0} & \textbf{9.1} \\
IID replay & 51.7 & 75.6 & 2.4 & 7.4 \\
Elite SFT$\rightarrow$GRPO & 56.9 & 68.5 & 1.7 & 3.0 \\
\bottomrule
\end{tabular}
\caption{\textbf{Ablations of the archive and of replay during RL} on BIRD. IID replay replaces the elite archive with verified samples from the base model, and Elite SFT$\rightarrow$GRPO fine-tunes on the elite archive before GRPO, without replay. \method{} produces the most distinct correct strategies.}
\label{tab:ablations}
\end{table}

\paragraph{Preserving diversity requires replay throughout RL.}
To test whether an SFT warm-up is sufficient, we train with GRPO after fine-tuning on the elite archive, without further replay (\emph{Elite SFT$\rightarrow$GRPO}). As shown in Table~\ref{tab:ablations}, both pass@$128$ and U@$128$ fall to levels comparable to standard GRPO (68.5\% and 3.0 versus 67.9\% and 3.0). In contrast, \method{} maintains high coverage and diversity by continuing to replay the archive during RL. Thus, exposure to diverse solutions before training is insufficient to prevent their subsequent collapse, consistent with findings that integrating off-policy data into RL is more effective than a separate SFT stage \citep{yan2025luffy,zhang2025chord}.

\paragraph{Archive composition matters even when replay is used.}
Finally, we replace the elite archive with IID samples while keeping the replay objective unchanged (\emph{IID replay}). Although IID replay achieves comparable pass@$k$, it produces fewer distinct correct strategies at every sampling budget, reaching U@$128$ of 7.4 compared to 9.1 for \method{} (Table~\ref{tab:ablations}). Because IID samples overrepresent the base model's frequent approaches, replaying them offers less retention of the rarer strategies that RL suppresses. These ablations show that preserving diversity depends on both constructing an archive with broad strategy coverage and continuing to replay it throughout RL.
\section{Related Work}
\label{sec:related}

\paragraph{Diversity in RL post-training.} RL post-training is widely observed to reduce output diversity, with studies showing that it lowers diversity relative to supervised fine-tuning \citep{kirk2024understanding}, collapses policy entropy \citep{cui2025entropy}, and concentrates probability on behaviors already present in the base model \citep{zhao2025echo}, so that base models often solve more problems at large sampling budgets \citep{yue2025does,oh2026sharpening}; supervised fine-tuning shows a similar collapse \citep{dang2025weight}. Mitigations optimize objectives tied to pass@$k$ \citep{walder2025pkpo,chen2025passk,tajwar2026maxrl}, reward novel or diverse outputs \citep{song2025outcome,li2025darling,li2026setpo,jiang2026mentor}, regularize toward the base model \citep{li2025dphrl}, or control entropy and sampling temperature \citep{cui2025entropy,yu2025dapo,oh2026sharpening}, and are typically evaluated by pass@$k$ \citep{yu2025passk}. 
\method{} instead replays an archive built for coverage, and we measure diversity at the more meaningful level of solution strategies.

\paragraph{Mixing demonstrations and replay into RL.} RL has long been combined with supervised learning from demonstrations or past experience \citep{lin1992self,hester2018dqfd,oh2018sil}. In LLM post-training, the off-policy data range from pretraining text and expert solutions \citep{ouyang2022instructgpt,zhang2025chord,yan2025luffy} to replayed verified trajectories \citep{zhang2025rlep,wang2025eframe} and base-model samples \citep{li2025dphrl}. These works use replay to learn efficiently from good solutions or to stay close to a reference model. \method{} uses a replay set built by quality-diversity search to keep distinct correct solutions available throughout RL.

\paragraph{Quality diversity and evolutionary RL.} QD algorithms such as novelty search and MAP-Elites seek collections of solutions that are both high-performing and diverse \citep{lehman2011abandoning,mouret2015illuminating,pugh2016quality,cully2018quality}. In deep RL, they have been combined with policy gradients through shared replay buffers \citep{khadka2018evolution,pourchot2019cemrl}, gradient-based variation \citep{nilsson2021pga,pierrot2022qdpg}, and archive-driven exploration \citep{ecoffet2021first}.
In these works, QD discovers new behaviors, and archived solutions come from a separate search process. \method{} instead builds its archive once from the base model's own solutions and replays it to preserve its diversity of behavior.

\section{Conclusion}
\label{sec:conclusion}

We introduced \method{}, which preserves solution diversity during RL post-training by replaying an archive of distinct correct solutions found by quality-diversity search. \method{} raises accuracy both in and out of distribution while retaining most of the base model's correct strategies. We also introduced a strategy-level evaluation of diversity, which reveals differences between methods that pass@$k$ conceals. Our results show that post-training need not trade diversity for reliability.
\section*{Acknowledgments}
We are grateful to Amazon Research Award, Ohio Supercomputing Center (OSC), and NAIRR Pilot Program (NAIRR250264) for providing the compute resources for this work. HNM is partially supported by the Center on Responsible AI and Governance (CRAIG) at OSU.

\section*{Limitations}
We recognize the following limitations and directions of future work. 

\paragraph{Base model.}
Our experiments use Qwen3.5-4B-Base, whose moderate size makes it feasible to conduct multiple training runs across our method and baselines within our available compute budget. This model size is also relevant to practical RL post-training settings, making it a meaningful model for studying strategy preservation. Nevertheless, evaluating \method{} on larger models (e.g. 8B), and across different model families is a natural direction for future work.

\paragraph{Domain.}
Our experiments focus on text-to-SQL, which provides a suitable testbed for studying solution diversity because questions can admit multiple correct queries, correctness can be verified through execution, and SQL's formal structure allows strategies to be distinguished using abstract syntax trees. Extending this evaluation to other domains where solution strategies lack a clear structural representation, as in many mathematical reasoning tasks, might be challenging. Nevertheless, we expect the approach to be applicable to other domains with verifiable solutions and meaningful behavioral distinctions, such as program optimization or algorithmic coding and we view this as an important direction for future work.

\section*{AI Use Statement}
AI tools were used to assist with editing the manuscript and with writing the code used to run the experiments. All text, code, results, and claims were reviewed by the authors, who take full responsibility for the content of this work.
\bibliography{references}

\clearpage
\appendix
\section*{Appendix}
We provide details of \method{} method, (Appendix~\ref{app:method_details}), our text-to-SQL instantiation together with experimental and evaluation settings (Appendix~\ref{app:setup}), and additional analyses and examples (Appendix~\ref{app:results}).

\newtcblisting{appendixcode}[2][]{
  enhanced, breakable, listing only,
  colback=black!2, colframe=black!30, boxrule=0.45pt,
  arc=2pt, left=5pt, right=5pt, top=4pt, bottom=4pt,
  fonttitle=\bfseries\footnotesize, title={#2},
  listing options={basicstyle=\ttfamily\scriptsize, breaklines=true,
    breakatwhitespace=true, columns=fullflexible, keepspaces=true,
    showstringspaces=false}, #1
}

\section{Method Details}
\label{app:method_details}

\subsection{Algorithm}
\label{app:algorithm}
Algorithm~\ref{alg:method} summarizes the two stages of \method{}. We first construct an archive with the frozen base model, then use the archived solutions to guide GRPO post-training.

\begin{algorithm}[t]
\caption{\method{}}
\label{alg:method}
\begin{algorithmic}[1]
\Require base model $\pi_0$; training questions $\mathcal{D}$; archive questions $\mathcal{D}_A\subseteq\mathcal{D}$; verifier $r$; descriptor $b$; replay weight $\mu$
\Statex \textbf{Stage 1: construct the elite archive}
\State collect verified seed solutions from $\pi_0$ and induce an approach vocabulary
\For{each $x\in\mathcal{D}_A$}
  \State initialize $\mathcal{A}_x$ from verified seeds, one elite per cell $b(y)$
  \Repeat
    \State choose an applicable target approach and a parent elite
    \State generate a candidate $y'$ with $\pi_0$ conditioned on the target and parent
    \If{$r(x,y')=1$}
      \State assign $y'$ to its realized cell, replacing the elite if shorter
    \EndIf
  \Until{search budget exhausted}
\EndFor
\State select verified elites across questions to form the replay set
\Statex \textbf{Stage 2: train with elite replay}
\State $\pi_\theta\gets\pi_0$
\Repeat
  \State compute $\mathcal{L}_{\mathrm{GRPO}}$ from verifier-scored on-policy responses
  \State sample a batch $\mathcal{E}$ from the replay set
  \State $\mathcal{L}_{\mathrm{replay}}\gets -\frac{1}{|\mathcal{E}|}\sum_{(x,y)\in\mathcal{E}}\frac{1}{|y|}\log\pi_\theta(y\mid x)$
  \State update $\theta$ to minimize $(1-\mu)\mathcal{L}_{\mathrm{GRPO}}+\mu\mathcal{L}_{\mathrm{replay}}$
\Until{training budget exhausted}
\State \Return $\pi_\theta$
\end{algorithmic}
\end{algorithm}

\subsection{Behavior space}
\label{app:behavior}
We first sample verified seed solutions from the frozen base model and present representative examples to a separate LLM, the \emph{labeler}. Rather than restricting search to the approaches observed in the seeds, the labeler uses these examples to induce a broader vocabulary of possible approaches. It then assigns each verified solution a primary approach label. Because solutions with the same approach label can still differ in structure, we refine each label with structural features computed deterministically from the solution, such as which constructs appear in its parse tree. The resulting descriptor $b(y)$ identifies the solution's cell in its question-specific archive $\mathcal{A}_x$. Each cell contains at most one verified solution, or \emph{elite}. The descriptor only guides search; diversity is evaluated with a separate representation that involves no labeler (Appendix~\ref{app:descriptors}).

\subsection{Archive search}
\label{app:search}
Search runs on a subset $\mathcal{D}_A$ of the training questions; no evaluation questions are used. All candidates come from the frozen base model $\pi_0$, prompted with the original task, and only candidates accepted by the verifier enter the archive. Each question receives a number of initial samples, followed by several rounds of mutations.

\paragraph{Target selection.}
For each question, the labeler proposes a set of applicable approaches. For each mutation, we first restrict target selection to applicable approaches not yet represented by any elite for the current question; if all are represented, we consider all applicable approaches. Within this pool, we prioritize approaches with the fewest previous targeting attempts for that question, then those represented by the fewest questions across the archive.

\paragraph{Variation and insertion.}
For each attempt, the frozen base model receives the original task, the target approach, and a verified parent solution, preferably using a different approach. The generated candidate is verified and assigned to the cell corresponding to the approach it \emph{actually} uses, which might not be the requested one. A candidate fills an empty cell or replaces its current elite if shorter. After search, we select verified elites across questions to form the replay set, each paired with its original task prompt. Targeting unrepresented approaches first steers the base model toward approaches it rarely produces on its own, and counting previous attempts prevents search from repeatedly targeting an approach the base model fails to realize. A candidate that repeats an already occupied approach can at most replace that cell's elite, so near-duplicate samples consume search budget but do not add coverage. Appendices~\ref{app:archive-growth} and~\ref{app:iid} show that this procedure finds substantially more distinct approaches than direct sampling.

\subsection{Training with elite replay}
\label{app:replay}
Each training step pairs a batch of on-policy GRPO rollouts with a separate batch of verified elites for the replay loss. We sample the latter across questions and occupied cells so that questions with more archived solutions do not dominate. The replay loss averages each solution's negative log-likelihood over its tokens, then averages over solutions. Inspired by CHORD~\citep{zhang2025chord}, we combine this term with GRPO using a fixed weight $\mu$ rather than a scheduled weight or additional per-token reweighting. The archive remains fixed throughout training.

\section{Experimental Details}
\label{app:setup}
This section describes our instantiation of \method{} for text-to-SQL, followed by the training and evaluation settings used in all experiments.

\subsection{Archive construction for text-to-SQL}
\label{app:sql-archive}
\label{app:archive}

\paragraph{Questions and verification.}
We train on 5,000 BIRD questions and construct the archive using 1,000 of them, sampled proportionally across databases and difficulty levels. These questions remain in the RL training pool. All candidates come from the frozen Qwen3.5-4B-Base model, prompted with the question, database schema, and BIRD evidence. We retain only SQL queries that execute successfully and match the gold query's result on the corresponding database.

\paragraph{Behavior space.}
We present up to 1,000 representative verified seed queries to Qwen3-Coder-30B-A3B-Instruct, which serves as the labeler. The structural features combined with the approach label are extracted from the SQL parse, including joins, subqueries, aggregation, and the use of set operations or ordering with a limit. Table~\ref{tab:approaches} illustrates examples from the vocabulary. 

\begin{table}[t]
\centering\small
\begin{tabular}{@{}p{0.35\columnwidth}p{0.55\columnwidth}@{}}
\toprule
\textbf{Approach label} & \textbf{Typical construction} \\
\midrule
Join with filter & Join relations, then restrict rows \\
Nested query structure & Solve through multiple subqueries \\
Conditional aggregation & Aggregate only selected records \\
Existence check & Test whether related records exist \\
Group-by aggregate & Aggregate separately within groups \\
Sort and limit & Rank results and retain a subset \\
\bottomrule
\end{tabular}
\caption{\textbf{Examples from the induced approach vocabulary.} The full vocabulary contains 24 approaches. Archive cells additionally distinguish structural features within these broad labels.}
\label{tab:approaches}
\end{table}

\paragraph{Search and replay-set selection.}
Each question receives eight initial samples, followed by eight rounds of eight mutations, for 72 candidates per question. The labeler proposes up to eight applicable approaches per question. After search, we select verified elites across questions to form a replay set of 1,500 examples. Table~\ref{tab:archive} summarizes the resulting archive.

\begin{table}[t]
\centering\small
\begin{tabular}{@{}lr@{}}
\toprule
Questions searched & 1,000 \\
Candidates generated & 72,000 \\
Verified candidates & 38,139 \\
Occupied cells after search & 2,708 \\
\midrule
Replay examples & 1,500 \\
Questions represented & 780 \\
\bottomrule
\end{tabular}
\caption{\textbf{Elite archive statistics on BIRD.} Questions without a verified solution are not represented in the replay set.}
\label{tab:archive}
\end{table}

\subsection{Prompts for behavior-space construction and search}
\label{app:prompts}

We use a separate labeler to describe verified seed solutions, construct the
approach vocabulary, and assign approaches to candidates. The following are
the prompt templates. The labeler receives the system instruction below.

\begin{appendixcode}{Common labeler system instruction}
You are a strategy taxonomy annotator. Never solve the task or emit a replacement answer. Return only the requested JSON metadata. Describe computational/query structure, never wording or formatting.
\end{appendixcode}

For seed description and solution classification, the input includes the original task and its candidate query:

\begin{appendixcode}{Labeler input for describing and classifying a query}
Original task:
{Question, database schema, and supporting information}

Candidate SQL:
{SQL query to describe or classify}
\end{appendixcode}

\begin{appendixcode}{Describing a verified seed solution}
Describe the strategy used by this verified solution in one concise sentence. Return JSON as {"strategy_description": "..."}.

Problem id: {problem_id}
{labeler_input}
\end{appendixcode}

\begin{appendixcode}{Consolidating the approach taxonomy}
Given the following descriptions of verified SQL
solutions from BIRD, consolidate them into a set
of reusable labels representing distinct approaches
to solving SQL questions.

For each label, provide:
- A unique identifier and a descriptive name.
- A definition of the approach.
- Positive cues: features indicating that a query
  follows this approach.
- Negative cues: features distinguishing this
  approach from other, similar approaches.

Return a JSON object with the following structure:
{
  "labels": [
    {
      "id": "stable_snake_case",
      "name": "...",
      "definition": "...",
      "positive_cues": ["..."],
      "negative_cues": ["..."]
    }
  ]
}

Strategy descriptions:
- {description_1}
- {description_2}
...
\end{appendixcode}

\begin{appendixcode}{Classifying a verified solution}
Given a verified SQL query and a set of predefined
solution approaches, identify the approach that best
describes how the query solves the question.

Assign one primary approach label, and optionally a
secondary label if another approach is also relevant.
Use only labels from the provided taxonomy. Briefly
justify your choice and indicate your confidence.

Available approaches:
{approach labels and their definitions}

Original task:
{question, database schema, and supporting information}

Verified SQL query:
{SQL query}

Return:
{
  "primary_strategy": "id",
  "secondary_strategy": null,
  "confidence": 0.0,
  "justification": "..."
}
\end{appendixcode}

\begin{appendixcode}{Selecting applicable approaches}
Given a question and the available approach labels,
identify up to eight distinct approaches that could
plausibly produce a correct SQL solution.

Consider approaches beyond those already observed
in the archive to encourage exploration of alternative
ways to solve the question. Select only approaches
from the provided taxonomy, and return their labels
without generating any SQL queries.

Available approaches:
{approach labels and their definitions}

Original task:
{question, database schema, and supporting information}

Approaches already observed:
{labels of existing solutions}

Return:
{"strategy_ids": ["id", "..."]}
\end{appendixcode}

During search, the following instruction is appended to the original task prompt supplied to the frozen base model.

\begin{appendixcode}{Generating a candidate using a target approach}
Solve the original question using the target approach
described below. The solution must be correct and
substantially different in structure from the verified
parent solution, if provided.

You may use the parent as a starting point, but aim
to produce a distinct solution following the target
approach. Return only one executable SQL query,
without explanations or alternative solutions.

Target approach:
{approach name and definition}

Verified parent solution (if available):
{parent SQL query}
\end{appendixcode}

The labeler proposes applicable approaches, while the search procedure chooses the next target (Appendix~\ref{app:search}). The parent block in the mutation prompt is omitted when no verified parent is available.

\subsection{Implementation and training configuration}
\label{app:compute}
We implement training using veRL~\citep{sheng2025hybridflow} as the distributed RL framework, integrated with Trinity-RFT~\citep{pan2025trinity} to support the combination of on-policy GRPO updates with the supervised replay loss. We use vLLM~\citep{kwon2023vllm} for rollout generation. All model parameters are trained on four NVIDIA A100 80 GB GPUs, with two allocated to policy optimization and two to rollout generation. For a consistent comparison, all methods use the same BIRD training data and are trained for 100 steps, with the principal shared settings summarized in Table~\ref{tab:training-settings}. Each baseline follows its respective training objective, specifically:
\paragraph{MaxRL.}
MaxRL uses the same rollout groups and clipped policy objective
but replaces GRPO's standard-deviation normalization with
mean-reward normalization:
\[
A_i^{\mathrm{MaxRL}}
=
\frac{r_i-\bar r}{\bar r+\epsilon},
\qquad \epsilon=10^{-6}.
\]
These advantages are used directly, without a subsequent GRPO
standardization. MaxRL uses no replay or additional
reference-model KL penalty.

\paragraph{SetPO.}
We combine GRPO advantages with the authors' set-level
diversity marginals:
\[
A_i^{\mathrm{SetPO}}
=
A_i^{\mathrm{GRPO}}+\lambda_{\mathrm{set}}m_i,
\]
where $\lambda_{\mathrm{set}}=0.05$ and the marginal-scoring
parameter is $\tau=1.0$. Marginals are computed within each
eight-response group using embeddings from the frozen
Qwen3-Embedding-0.6B model.

\paragraph{DPH-F.}
We first sample eight responses from the frozen base model for
each training question. Questions for which all eight responses
are correct are eligible for preservation. We then draw an
independent reference response for each eligible question;
questions whose reference response is incorrect return to the
exploration partition. This procedure yields 301 preservation
questions and 4,699 exploration questions. Reference response
tokens and base-model token log probabilities are cached.

Training combines GRPO on the exploration partition with
forward-KL preservation on 128 cached reference responses per
iteration, using coefficient $\beta=0.01$. We retain the
released low-variance token-level KL estimator rather than
replacing it with supervised cross-entropy.

\paragraph{EliteReplay and IID replay.}
Both replay methods use 128 verified archive responses per
iteration, alongside 128 questions with eight on-policy responses each for GRPO, and optimize
\[
\mathcal{L}
=
(1-\mu)\mathcal{L}_{\mathrm{GRPO}}
+\mu\mathcal{L}_{\mathrm{replay}},
\qquad \mu=0.05
\]
(Appendix~\ref{app:mu}).
The supervised term averages token negative log likelihood
within each response and then averages across replay responses.
Replay cycles across questions and their occupied archive
modes. The IID control uses the same replay objective and
sampling procedure, but its archive is constructed from
independent base-model samples rather than directed
quality-diversity search. Both archives contain
1,500 verified responses.
\begin{table}[t]
\centering\small
\begin{tabular}{@{}p{0.59\columnwidth}p{0.30\columnwidth}@{}}
\toprule
\textbf{Setting} & \textbf{Value} \\
\midrule
Base model & Qwen3.5-4B-Base \\
Training steps & 100 \\
Questions / step & 128 \\
On-policy responses / question & 8 \\
Optimizer & AdamW \\
Learning rate & $10^{-6}$ \\
Policy clip interval & $[0.8,1.28]$ \\
Generation temperature / top-$p$ & 1.0 / 0.95 \\
Prompt / response limit & 16,384 / 4,096 tokens \\
Replay examples / step & 128 \\
Replay mixing weight & $\mu=0.05$ \\
\bottomrule
\end{tabular}
\caption{\textbf{Principal training settings.} All methods use the same training questions and sampling budget; objective-specific parameters vary by baseline.}
\label{tab:training-settings}
\end{table}

\subsection{Evaluation datasets and correctness checks}
\label{app:verification}
\paragraph{Text-to-SQL.}
We evaluate on BIRD dev (1,534 questions) and Spider 1.0 dev (1,034 questions). For each model and question, we draw $n=128$ samples using temperature 1.0 and top-$p$ 0.95. SQL correctness is determined by executing the generated and gold queries on the benchmark database and comparing their results. Invalid or non-executable queries are counted as incorrect. Execution accuracy establishes correctness on the given database, not equivalence on all possible databases.

\paragraph{MATH-500 and AmbiQT.}
MATH-500 tests generalization beyond SQL using 128 samples per problem and the same answer-checking procedure for all models. AmbiQT evaluates whether the sampled SQL queries cover both valid readings of an ambiguous question. Both@$k$ is the fraction of questions for which $k$ samples contain a correct query for each reading. We evaluate on a reproducible, proportionally stratified sample of 500 AmbiQT questions. Appendix~\ref{app:ambiqt} gives an example of a task from this benchmark.

\subsection{Strategy descriptors and evaluation diversity metrics}
\label{app:descriptors}
The search descriptor $b(y)$ combines an LLM-assigned approach label with structural SQL features, while evaluation relies on a separate, deterministic AST-based representation. This separation avoids evaluating diversity using the same labels that guide the search. The evaluation representation captures features such as joins, subqueries, aggregation, window functions, set operations, and ordering with a limit, distinguishing different solution structures while ignoring superficial differences such as table aliases.
\subsection{Retention and coverage change}
\label{app:retention}
\paragraph{Retention.}
For each question, we identify the strategies among the base model's correct samples and compute the fraction also represented among a post-trained model's correct samples. Each strategy counts once regardless of its frequency, and the fractions are averaged equally over questions. We report retention on 825 BIRD and 562 Spider questions where the base model produces at least two correct strategies.

\paragraph{Coverage change.}
For each common strategy, we compare the percentage of questions on which the post-trained model produces at least one correct query of that type with the corresponding percentage for the base model. BIRD coverage is computed over the 1,131 questions on which the base model produces a correct query. A decrease of 22 percentage points therefore means that the strategy appears on 22 points fewer questions. Less common strategies are grouped as ``other.''

\subsection{AmbiQT example}
\label{app:ambiqt}
The question ``Return the average earnings across all poker players'' admits two readings because the schema provides individual earnings and a precomputed average. Both of the following queries are valid interpretations.

\begin{appendixcode}{Two valid readings of an AmbiQT question}
-- Compute the mean from individual earnings
SELECT AVG(earnings) FROM poker_player;

-- Retrieve the precomputed mean
SELECT avg_earnings FROM poker_player_earnings;
\end{appendixcode}

Both@$k$ checks whether the sampled set covers both readings, rather than only one.

\subsection{Case study selection}
\label{app:cases}
The BIRD and Spider case studies are drawn from questions on which the base model produced several correct strategies and every method produced at least one correct sample. Among these, we chose examples spanning difficulty levels and approach types. They are meant to illustrate the per-question behavior behind the aggregate retention and coverage results (Figure~\ref{fig:retention}; Appendix~\ref{app:delta-spider}), which are computed over all eligible questions. Representative queries appear in Appendix~\ref{app:queries}.
\section{Additional Results}
\label{app:results}

\subsection{Archive growth during quality-diversity search}
\label{app:archive-growth}
MAP-Elites increases the number of occupied behavior cells from 1,132 after seed sampling to 2,708 after eight mutation rounds (Table~\ref{tab:archive-growth}). Thus, targeted search finds verified approaches beyond those obtained by initial sampling alone. 
\begin{table}[t]
\centering\small
\begin{tabular}{@{}lrr@{}}
\toprule
\textbf{Stage} & \textbf{Candidates} & \textbf{Occupied cells} \\
\midrule
Seeds & 8,000 & 1,132 \\
After round 2 & 24,000 & 1,744 \\
After round 4 & 40,000 & 2,125 \\
After round 8 & 72,000 & 2,708 \\
\bottomrule
\end{tabular}
\caption{\textbf{Growth of verified behavior coverage during archive search.}}
\label{tab:archive-growth}
\end{table}

\subsection{Comparing elite and IID archives}
\label{app:iid}
The IID comparison archive is generated by sampling directly from the same 1,000 questions, without targeted mutations. Both archives use 72,000 generation attempts and supply 1,500 verified replay examples. This isolates the benefit of targeted search from simply increasing the number of sampled candidates. Among 547 questions represented by at least two solutions in both archives, pairs sampled from the elite archive exhibit different strategies (based on AST similarity) more often than those from the IID archive (66.2\% versus 45.5\%).


\subsection{Replay weight}
\label{app:mu}
Table~\ref{tab:mu} varies the replay weight $\mu$ on BIRD. Larger weights modestly reduce pass@$1$, whereas pass@$128$ for all settings remains in a similar range.

\begin{table}[h]
\centering
\small
\begin{tabular}{l cccc}
\toprule
 & p@1 & p@8 & p@32 & p@128 \\
\midrule
$\mu=0.02$ & 53.4 & 66.2 & 70.2 & 73.1 \\
$\mu=0.05$$^\ast$ & 52.5 & 66.2 & 71.0 & 74.8 \\
$\mu=0.10$ & 51.0 & 65.8 & 71.0 & 74.3 \\
$\mu=0.15$ & 50.3 & 65.7 & 70.8 & 74.0 \\
\bottomrule
\end{tabular}
\caption{\textbf{Effect of the replay weight $\mu$} on BIRD (pass@$k$, \%). $^\ast$Default used in all other experiments.}
\label{tab:mu}
\end{table}

\subsection{Coverage change on BIRD and Spider}
\label{app:delta-spider}
\begin{figure*}[h]
\centering
\includegraphics[width=\textwidth]{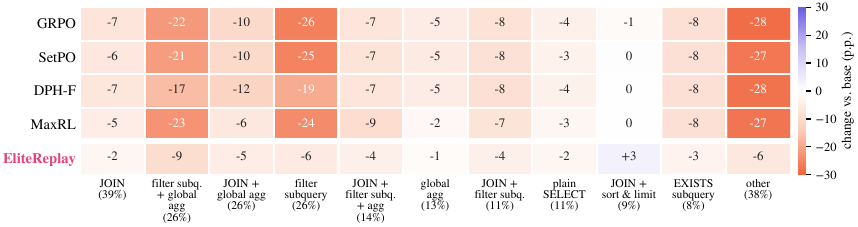}
\caption{\textbf{Change in strategy coverage relative to the base model on BIRD}, in percentage points of questions; column labels give the base model's coverage. \method{} loses less than every baseline on every strategy.}
\label{fig:delta}
\end{figure*}

\begin{figure*}[h]
\centering
\includegraphics[width=\textwidth]{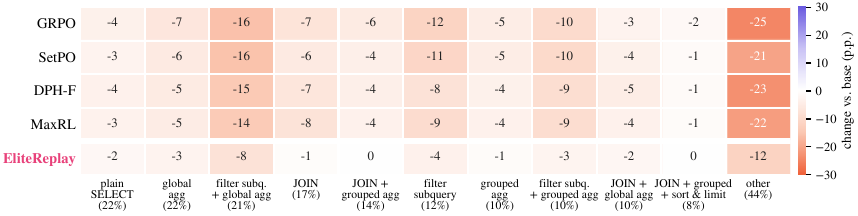}
\caption{\textbf{Change in strategy coverage relative to the base model on Spider.} \method{}'s losses are again the smallest on every strategy.}
\label{fig:delta-spider}
\end{figure*}

Figure~\ref{fig:delta} shows the change in per-strategy coverage on BIRD discussed in \S\ref{sec:results-retention}, and Figure~\ref{fig:delta-spider} repeats the strategy-coverage analysis on Spider. As on BIRD, \method{} loses less coverage than the baselines across the base model's common strategies.


\subsection{Case studies on Spider}
\label{app:cases-spider}
\begin{figure*}[h]
\centering
\includegraphics[width=\textwidth]{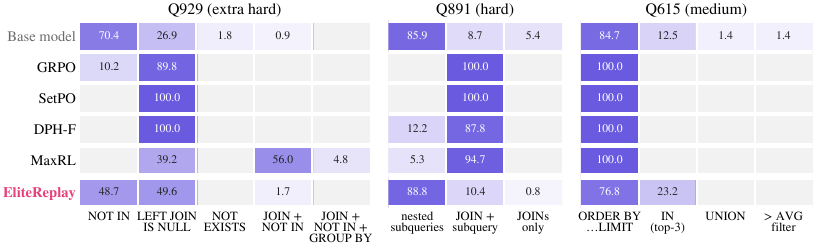}
\caption{\textbf{Distribution of correct samples over strategies on three Spider questions}, in the format of Figure~\ref{fig:cases}.}
\label{fig:cases-app}
\end{figure*}

Figure~\ref{fig:cases-app} shows three Spider questions in the same format as Figure~\ref{fig:cases}. On Q929, for example, \method{} preserves both the \texttt{NOT IN} and \texttt{LEFT JOIN ... IS NULL} approaches, whereas several baselines concentrate on only one.

\subsection{Representative queries}
\label{app:queries}

We provide one representative SQL query for each strategy shown in the BIRD and Spider case studies (Figures~\ref{fig:cases} and~\ref{fig:cases-app}).

\paragraph{BIRD Q212 (challenging).}
``Which element is the least numerous in non-carcinogenic molecules?''

\begin{appendixcode}{Join, membership-subquery, and derived-table approaches}
-- JOIN
SELECT atom.element
FROM atom
JOIN molecule
  ON atom.molecule_id = molecule.molecule_id
WHERE molecule.label = '-'
GROUP BY atom.element
ORDER BY COUNT(*) ASC LIMIT 1;

-- IN subquery
SELECT element
FROM atom
WHERE molecule_id IN (
  SELECT molecule_id FROM molecule
  WHERE label = '-'
)
GROUP BY element
ORDER BY COUNT(*) ASC LIMIT 1;

-- Derived table
SELECT element
FROM atom
JOIN connected
  ON atom.atom_id = connected.atom_id
JOIN molecule
  ON atom.molecule_id = molecule.molecule_id
WHERE molecule.label = '-'
GROUP BY element
HAVING COUNT(element) = (
  SELECT MIN(element_count)
  FROM (
    SELECT COUNT(element) AS element_count
    FROM atom
    JOIN connected
      ON atom.atom_id = connected.atom_id
    JOIN molecule
      ON atom.molecule_id = molecule.molecule_id
    WHERE molecule.label = '-'
    GROUP BY element
  )
);
\end{appendixcode}

\paragraph{BIRD Q1146 (moderate).}
``Please provide the full name of the away team that scored the most goals.''

\begin{appendixcode}{Alternative ways to identify the maximum}
-- MAX() filter
SELECT Team.team_long_name
FROM Team
JOIN "Match"
  ON Team.team_api_id = "Match".away_team_api_id
WHERE "Match".away_team_goal = (
  SELECT MAX(away_team_goal) FROM "Match"
);

-- Grouped ranking
SELECT team_long_name
FROM Team
WHERE team_api_id = (
  SELECT away_team_api_id
  FROM Match
  GROUP BY away_team_api_id
  ORDER BY MAX(away_team_goal) DESC
  LIMIT 1
);

-- MAX() + LIMIT
SELECT t.team_long_name
FROM Team t
JOIN Match m
  ON t.team_api_id = m.away_team_api_id
WHERE m.away_team_goal = (
  SELECT MAX(m2.away_team_goal) FROM Match m2
)
ORDER BY m.away_team_goal DESC LIMIT 1;

-- ORDER BY ... LIMIT
SELECT t.team_long_name
FROM "Team" t
JOIN "Match" m
  ON m.away_team_api_id = t.team_api_id
ORDER BY m.away_team_goal DESC LIMIT 1;
\end{appendixcode}

\paragraph{BIRD Q1110 (moderate).}
``Tell the build Up play passing class for "FC Lorient" on 2010/2/22.''

\begin{appendixcode}{Subquery and join approaches}
-- Subquery lookup
SELECT buildUpPlayPassingClass
FROM Team_Attributes
WHERE team_fifa_api_id = (
  SELECT team_fifa_api_id FROM Team
  WHERE team_long_name = 'FC Lorient'
)
AND date LIKE '2010-02-22%';

-- JOIN
SELECT `Team_Attributes`.`buildUpPlayPassingClass`
FROM `Team_Attributes`
JOIN `Team`
  ON `Team_Attributes`.`team_fifa_api_id`
   = `Team`.`team_fifa_api_id`
WHERE `Team`.`team_long_name` = 'FC Lorient'
AND `Team_Attributes`.`date` LIKE '2010-02-22%';

-- Subquery + LIMIT
SELECT "buildUpPlayPassingClass"
FROM Team_Attributes
WHERE "team_api_id" = (
  SELECT "team_api_id" FROM Team
  WHERE "team_long_name" = 'FC Lorient'
)
AND "date" LIKE '2010-02-22%'
ORDER BY "date" DESC LIMIT 1;
\end{appendixcode}

\paragraph{Spider Q929 (extra hard).}
``Give me the id, role and email of the professionals who did not perform any treatment on dogs.''

\begin{appendixcode}{Anti-join approaches and structural variants}
-- NOT IN
SELECT Professionals.professional_id,
       Professionals.role_code,
       Professionals.email_address
FROM Professionals
WHERE Professionals.professional_id NOT IN (
  SELECT Treatments.professional_id
  FROM Treatments
);

-- LEFT JOIN ... IS NULL
SELECT Professionals.professional_id,
       Professionals.role_code,
       Professionals.email_address
FROM Professionals
LEFT JOIN Treatments
  ON Professionals.professional_id
   = Treatments.professional_id
WHERE Treatments.treatment_id IS NULL;

-- NOT EXISTS
SELECT P.professional_id, P.role_code,
       P.email_address
FROM Professionals P
WHERE NOT EXISTS (
  SELECT 1 FROM Treatments T
  WHERE T.professional_id = P.professional_id
);

-- JOIN + NOT IN
SELECT Professionals.professional_id,
       Professionals.role_code,
       Professionals.email_address
FROM Professionals
LEFT JOIN Treatments
  ON Professionals.professional_id
   = Treatments.professional_id
WHERE Professionals.professional_id NOT IN (
  SELECT Treatments.professional_id FROM Treatments
);

-- JOIN + NOT IN + GROUP BY
SELECT Professionals.professional_id,
       Professionals.role_code,
       Professionals.email_address
FROM Professionals
LEFT JOIN Treatments
  ON Professionals.professional_id
   = Treatments.professional_id
WHERE Professionals.professional_id NOT IN (
  SELECT Treatments.professional_id
  FROM Treatments
  GROUP BY Treatments.professional_id
  HAVING COUNT(*) > 0
);
\end{appendixcode}

The final two forms add structural elements around \texttt{NOT IN}, rather than representing entirely different anti-join operations.

\paragraph{Spider Q891 (hard).}
``Return the names of friends of the high school student Kyle.''

\begin{appendixcode}{Nested subqueries, join with subquery, and joins}
-- Nested subqueries
SELECT name
FROM Highschooler
WHERE ID IN (
  SELECT friend_id FROM Friend
  WHERE student_id = (
    SELECT ID FROM Highschooler
    WHERE name = 'Kyle'
  )
);

-- JOIN + subquery
SELECT Highschooler.name
FROM Highschooler
JOIN Friend
  ON Highschooler.ID = Friend.friend_id
WHERE Friend.student_id = (
  SELECT ID FROM Highschooler
  WHERE name = 'Kyle'
);

-- JOINs only
SELECT DISTINCT h2.name
FROM Highschooler h1
JOIN Friend f ON h1.ID = f.student_id
JOIN Highschooler h2 ON f.friend_id = h2.ID
WHERE h1.name = 'Kyle';
\end{appendixcode}

\paragraph{Spider Q615 (medium).}
``List top 3 highest Rating TV series. List the TV series's Episode and Rating.''

\begin{appendixcode}{Ranking, membership, set operation, and filtering}
-- ORDER BY ... LIMIT
SELECT Episode, Rating
FROM TV_series
ORDER BY Rating DESC LIMIT 3;

-- IN (top-3)
SELECT Episode, Rating
FROM TV_series
WHERE Rating IN (
  SELECT Rating FROM TV_series
  ORDER BY Rating DESC LIMIT 3
);

-- UNION
SELECT Episode, Rating FROM TV_series
WHERE Rating = (
  SELECT Rating FROM TV_series
  ORDER BY Rating DESC LIMIT 1
)
UNION
SELECT Episode, Rating FROM TV_series
WHERE Rating = (
  SELECT Rating FROM TV_series
  ORDER BY Rating DESC LIMIT 1 OFFSET 1
)
UNION
SELECT Episode, Rating FROM TV_series
WHERE Rating = (
  SELECT Rating FROM TV_series
  ORDER BY Rating DESC LIMIT 1 OFFSET 2
)
LIMIT 3;

-- Above-average filter
SELECT Episode, Rating
FROM TV_series
WHERE Rating > (
  SELECT AVG(Rating) FROM TV_series
)
ORDER BY Rating DESC LIMIT 3;
\end{appendixcode}
\end{document}